\documentclass[runningheads]{llncs}
\usepackage{verbatim}
\usepackage{orcidlink}
\usepackage{tabularx}
\usepackage{changepage}
\usepackage{hyperref}
\usepackage{arydshln} 
\usepackage[T1]{fontenc}
\usepackage{multirow}
\usepackage{graphicx}
\usepackage{amsmath}
\begin{document}

\title{ClinLinker: Medical Entity Linking of Clinical Concept Mentions in Spanish}
\titlerunning{ClinLinker: MEL in Spanish}
%
\author{Fernando Gallego\inst{1,2}*\orcidlink{0000-0001-8053-5707}
\and Guillermo López-García\inst{1,2}\orcidlink{0000-0001-5903-1483}
\and Luis Gasco-Sánchez\inst{3}\orcidlink{0000-0002-4976-9879}
\and Martin Krallinger\inst{3}\orcidlink{0000-0002-2646-8782}
\and Francisco J. Veredas\inst{1,2}\orcidlink{0000-0003-0739-2505}}
\authorrunning{F. Gallego et al.}
%
\institute{Departamento de Lenguajes y Ciencias de la Computaci\'on, Universidad de M\'alaga, M\'alaga, Spain \and Research Institute of Multilingual Language Technologies, Universidad de M\'alaga, M\'alaga, Spain \and Barcelona Supercomputing Center, Barcelona, Spain\\ \email{*fgallegodonoso@uma.es}}
%
\maketitle              
\begin{abstract}
Advances in natural language processing techniques, such as named entity recognition and normalization to widely used standardized terminologies like UMLS or SNOMED-CT, along with the digitalization of electronic health records, have significantly advanced clinical text analysis. This study presents ClinLinker, a novel approach employing a two-phase pipeline for medical entity linking that leverages the potential of in-domain adapted language models for biomedical text mining: initial candidate retrieval using a SapBERT-based bi-encoder and subsequent re-ranking with a cross-encoder, trained by following a contrastive-learning strategy to be tailored to medical concepts in Spanish. This methodology, focused initially on content in Spanish, substantially outperforming multilingual language models designed for the same purpose. This is true even for complex scenarios involving heterogeneous medical terminologies and being trained on a subset of the original data. Our results, evaluated using top-k accuracy at 25 and other top-k metrics, demonstrate our approach's  performance on two distinct clinical entity linking Gold Standard corpora, DisTEMIST (diseases) and MedProcNER (clinical procedures), outperforming previous benchmarks by 40 points in DisTEMIST and 43 points in MedProcNER, both normalized to SNOMED-CT codes. These findings highlight our approach's ability to address language-specific nuances and set a new benchmark in entity linking, offering a potent tool for enhancing the utility of digital medical records. The resulting system is of practical value, both for large scale automatic generation of structured data derived from clinical records, as well as for exhaustive extraction and harmonization of predefined clinical variables of interest. 
\keywords{Encoder-only large language model \and Contrastive learning \and Biomedical text mining \and Medical entity linking \and SNOMED-CT}
\end{abstract}


\begin{credits}
\subsubsection{\ackname} The authors acknowledge the support from the Spanish Ministerio de Ciencia e Innovación (MICINN) under projects PID2020-116898RB-I00, PID2020-119266RA-I00 and BARITONE (TED2021-129974B-C22).This work is also supported by the European Union’s Horizon Europe Co-ordination \& Support Action under Grant Agreement No 101080430 (AI4HF), Grant Agreement No 101058779 (BIOMATDB) as well as Grant Agreement No 101057849 (DataTool4Heartproject).

\subsubsection{\discintname}
The authors have no competing interests to declare that are relevant to the content of this article.
\end{credits}


\section{Introduction}

In the medical domain, significant advancements in natural language processing (NLP) have proven essential for the effective analysis of medical texts. The integration of NLP in healthcare has opened new avenues for patient care and research, allowing for more efficient and accurate analysis of large volumes of patient data. This progress is not just technological but also represents a paradigm shift in how medical data is processed and understood. Techniques such as named entity recognition (NER) and entity linking (EL), also known as entity normalization in the clinical NLP field, play a crucial role in the interpretation and utilization of electronic health records (EHR). NER, for instance, enables the identification of critical medical terms or clinical variables within unstructured data, while EL facilitate the mapping, normalization or harmonization of these terms to standardized medical controlled vocabularies. This process is vital for ensuring consistency and accuracy in data interpretation, as well as enabling data standardization, harmonization and semantic interoperability.

These technological advances align with the ongoing digital transformation in healthcare, highlighting the growing importance of EHRs in clinical practice. EHRs are more than digital versions of patient charts; they are comprehensive, interactive records that offer a holistic view of a patient's medical history. Their adoption has been pivotal in improving patient outcomes and healthcare efficiency. EHRs are invaluable to medical informatics due to their ability to consolidate diverse patient data, transforming unstructured data into a structured, analyzable knowledge base (KB). The richness of data in EHRs, including everything from clinical notes to laboratory results, presents both an opportunity and a challenge for NLP applications. The efficient extraction and analysis of this data have significant implications for clinical decision-making and patient care. However, integrating EHRs with NLP presents several challenges that must be addressed. One of these is the need for these systems to be adaptable and scalable to various healthcare settings and requirements.

\begin{figure}[!ht]
    \centering
        \includegraphics[width=\linewidth]{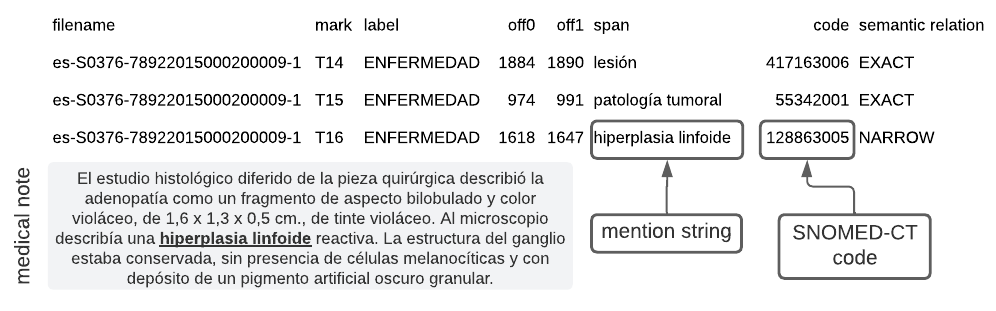}
    \caption{DisTEMIST-linking subtrack: requires automatically finding disease mentions in published clinical cases and assigning, to each mention, a SNOMED-CT term.}
    \label{fig:distemist_linking}
\end{figure}

The primary challenge for medical entity linking (MEL) involves handling heterogeneous mentions, where a controlled vocabulary concept is mentioned in practice through a diversity of written expressions or phrases. This heterogeneity can lead to misinterpretations and inconsistencies in data analysis, posing a risk to patient safety and care quality. Another important challenge arises when the mentions in the text do not have an exact match or have only a partial correspondence with the terms present in the KBs. This issue is particularly prevalent in free-text clinical notes, where the context and nuances of language can greatly affect meaning. It is also pivotal to account for variations, identifying multiple valid mentions for a single medical entity. These variations might include synonyms, acronyms, and different linguistic expressions, further complicating the text mining process. An overarching issue is data scarcity, particularly the lack of adequately annotated public medical records, being most of the current resources limited exclusively to content in English, thus making it even more urgent to generate resources and tools also for other languages. Moreover, the limited availability of datasets in less commonly spoken languages exacerbates the challenge of developing truly global NLP solutions. Figure~\ref{fig:distemist_linking} presents a MEL example from the DisTEMIST shared task~\cite{distemist}, illustrating disease entities linked to concepts from the SNOMED-CT terminology.

This backdrop motivates many recent developments in MEL, focusing on embedding similarity rather than conventional classification problems. Embedding-based approaches, leveraging deep learning (DL) techniques, have shown promise in capturing the semantic relationships between medical terms, enhancing the model's ability to understand and process complex medical language. Recent research has introduced multilingual language models designed as generic solutions for such challenges. These multilingual models are not only innovative from a technical point of view, but are also essential for bridging the language gap in medical informatics. Examples of these models include SapBERT~\cite{Liu2021-ik}, which starts from BERT-based language models---such as PubMedBERT~\cite{Gu2021-oo}, BioBERT~\cite{Lee2020-gp}, ClinicalBERT~\cite{Alsentzer2019-xu} or XLM-RoBERTa~\cite{Conneau2019-tz}, pre-trained with generic or domain-specific corpora, either monolingual or multilingual---, and continues with the pre-training of these models following a contrastive-learning strategy that allows self-aligning the spatial representation of biomedical entities, thus reaching the state-of-the-art (SOTA) on various MEL benchmarks. For this purpose, the authors used the entire Unified Medical Language System (UMLS) database\footnote{Unified Medical Language System\textsuperscript\textregistered (UMLS\textsuperscript\textregistered): \url{https://www.nlm.nih.gov/research/umls/knowledge_sources/metathesaurus/index.html}}, a meta-thesaurus of +4M medical concepts, to propose a new scalable learning framework for MEL task-focused models. SapBERT's ability to comprehend and interpret medical terminology across different languages represents a significant step forward in the field. 

Other studies aim to tailor these models to specific languages or standards using dedicated KBs. Such customization is essential for ensuring that the models are accurately aligned with local medical practices and terminologies. Recently, a number of evaluation campaigns have emerged that seek to encourage the development of approaches to specific corpora, such as DisTEMIST (see Figure~\ref{fig:distemist_linking}) and MedProcNER, focused on the detection and normalization of disease and procedure mentions in medical texts, respectively. These targeted approaches, while effective in specific contexts, highlight the need for more versatile and comprehensive solutions that can adapt to a broader range of medical texts and terminologies.

The UMLS Metathesaurus offers vocabularies in multiple languages, with Spanish having the second highest number of resources after English, providing extensive exploitation opportunities. This diversity in languages is key to developing NLP systems that are truly inclusive and globally applicable. Consequently, in this study we propose some MEL models, based on the SapBERT contrastive-learning strategy for self-alignment of medical concepts, which have trained solely on Spanish mentions from the entire UMLS (+1M medical concepts), including SNOMED-CT\footnote{SNOMED Clinical Terms\textsuperscript\textregistered (SNOMED-CT\textsuperscript\textregistered): \url{http://www.snomed.org/snomed-ct}}, ICD-10, and other terminological resources present in the metathesaurus, aiming to outperform generic language models on Spanish corpora and advance the SOTA. Our approach, focusing on a specific language edition, paves the way for more tailored and effective biomedical text mining applications in diverse linguistic environments.

\section{Related Work}

In the past decade, MEL has seen significant growth, primarily driven by the development of Transformer-based algorithms and increased computational capacity enabling their deployment. Additionally, various approaches have been proposed for obtaining valid candidates to map given entity mentions to standards like UMLS or SNOMED-CT. Among these methods, those ones that use KBs to enrich biomedical corpora win domain-specific information have shown promising results for MEL~\cite{lai2023keblm}. This strategy aims to surpass the limitations of conventional language models by incorporating structured data from biomedical KBs. The main advantage of these approaches is their ability to comprehend and process the complexities and nuances of medical texts. Techniques include adapting pre-trained language models and using contrastive learning objectives for mention encoding, improving not only machine learning (ML) convergence but also adaptability and scalability. However, a major limitation of these strategies is accessing high-quality KBs, which are often inaccessible, biased, or lacking in content to provide models with essential knowledge. In our work, this information is typically provided by a corpus-specific gazetteer and, in some cases, combined with other corpora from generic domains.

Another trend gaining prominence in recent years is combining techniques like prompt learning and generative artificial intelligence (AI). The impressive performance of these generative large language models (LLMs) has led to their adoption in biomedicine. The work of Yuan and his collaborators~\cite{yuan2022biobart} details the development of one of the first generative models for the biomedical domain, adapting the traditional BART model proposed by Lewis et al.~\cite{lewis2019bart} with PubMed abstracts, showing significant performance improvement in multiple language generation tasks. An advancement in this area is using prompt learning to fine-tune these models' responses. Thus, Ding et al.~\cite{ding-etal-2022-prompt} employ this method to enhance EL, especially in scenarios with few or zero samples (few- or zero-shot scenarios). For its part, the work of Yuan et al.~\cite{yuan2022generative} demonstrates the appropriateness of combining these two approaches, proposing training a generative model with a KB and creating synthetic samples with synonyms and definitions. The limitation of these approaches is, on the one hand, the need for high quality prompts and, on the other hand, the insufficient performance of these models in zero-shot inference setups. To date, there is not enough information on prompt optimization techniques to determine if this line of work is most suitable.

The predominant trend in which ClinLinker is focused is a two-phase approach: in the first stage a model generates the space with concept candidates for each entity mention; in the second stage another model selects the most relevant concepts/codes among the candidates by re-ranking them in order of their relevance. Some of-the-shell approaches use candidate generation based on a bi-encoder model and the subsequent re-ranking with a cross-encoder model. The work of Yu and collaborators~\cite{yu2021cross} emphasizes the importance of using cross-lingual models for improving long-text information retrieval and highlights the significance of employing specific pretraining and language adaptation for information retrieval tasks. A key advantage of our proposed approach over these methods is that, with an international version of the UMLS standard available, cross-lingual adaptation is unnecessary as the information is already contextually adapted. Additionally, techniques like cross-lingual typically rely on mapping more generic words rather than biomedical domain-specific ones. Furthermore, the training conducted on the cross-encoder proposed in this paper, based on hard triplets generated by the bi-encoder, enables the model to differentiate the closest optimal concept among more complex candidates, improving results both generally, with the entire corpus, and with previously unseen codes.

\section{Methods}
\label{sec:methods}

The following sections describe the corpora used for the training and evaluation of the MEL models, as well as the methodology followed for the MEL approach developed in this article.

\subsection{Description of the Corpora}
\label{sec:corpora}

For the DisTEMIST-linking subtrack~\cite{distemist} the organizers supplied the participants with a training dataset consisting of 750 annotated clinical cases. They also supplied a test dataset consisting of 250 unannotated clinical cases, together with a larger collection of 2,800 ``background'' clinical cases, to avoid manual corrections. For an example of an annotated document, see Figure~\ref{fig:distemist_linking}. The MedProcNER~\cite{LimaLpez2023OverviewOM} corpus is a collection of 1,000 clinical cases in Spanish from different medical specialties such as cardiology, oncology, otorhinolaryngology, dentistry, pediatrics, primary care, allergology, radiology, psychiatry, ophthalmology, and urology annotated with clinical procedure mentions. Every mention both in the MedProcNER and DisTEMIST corpus has been normalized using SNOMED-CT terminology.

\subsection{ClinLinker: Bi-encoder+Cross-encoder Pipeline for MEL}
\label{sec:pipeline}

\begin{figure}[!ht]
    \centering
        \includegraphics[width=\linewidth]{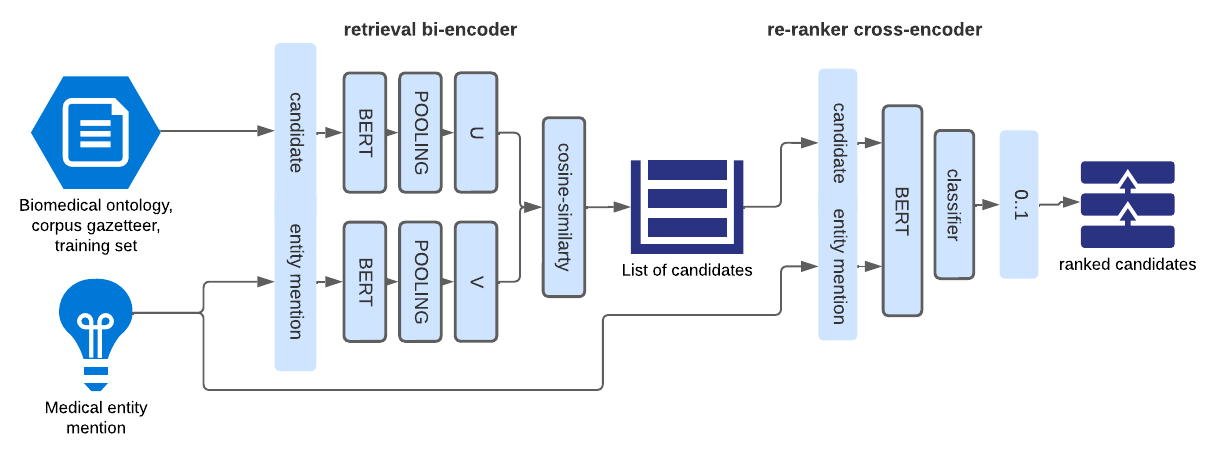}
    \caption{ClinLinker's two-stage pipeline for MEL: a first stage of candidate retrieval, using a bi-encoder, and subsequent stage of re-ranking, employing a cross-encoder.}
    \label{fig:bi_cross_pipeline}
\end{figure}

Our approach, which we call ClinLinker, consists of a two-stage pipeline for MEL of medical texts in Spanish: a first stage of candidate retrieval, using a SapBERT bi-encoder, and subsequent stage of re-ranking, employing a SapBERT cross-encoder (see Figure~\ref{fig:bi_cross_pipeline}). This approach addresses an EL problem by leveraging the similarity between the input mention and each candidate given by a bi-encoder previously trained. Our main contribution in this study is training the bi-encoder by using only medical concepts/codes (CUI\footnote{In UMLS, CUI is the Concept Unique Identifier.}) in Spanish from the UMLS corpus (+1M medical concepts), with two different approaches: 1) excluding (Spanish-SapBERT) and 2) including obsolete concepts/codes from the latest 2023 version (Spanish-SapBERT-oc). These comprehensive approaches ensure that our model is well-versed in both current and historical medical terminologies, enhancing its applicability in a wide range of clinical scenarios. To train a bi-encoder this way, once the entire Spanish UMLS corpus is pre-processed, for each CUI the full standard name (FSN) is used as an entity mention and every description or synonym of the CUI are used as positive candidates. This methodology aligns with the original scenario proposed in the SapBERT-model paper~\cite{Liu2021-ik}, a pretraining scheme that self-aligns the representation space of biomedical entities. Like SapBERT, we utilize \textit{multi-similarity-loss} (from the \textit{Pytorch Metrics Learning} library\footnote{https://kevinmusgrave.github.io/pytorch-metric-learning/}) as the loss function, so that for each \textit{(mention, positive candidate)} pair ($M_i$, $C_i$), the negative candidates concepts/codes are sourced from the same training batch using a hard-triplet mining strategy. The incorporation of negative candidates during the training of the bi-encoder---thus following what is called a contrastive-learning approach---sourced by using the model's own predictions, is critical as it teaches the model to discern between candidates that are close to but not exactly the correct match, a key aspect in fine-tuning its accuracy for MEL. The model chosen as the basis for the bi-encoder in this study is \textit{roberta-base-biomedical-clinical-es}~\cite{carrino2021biomedical}, a Transformer language model based on RoBERTa~\cite{Liu2019-yz} and chosen in this work for having been pre-trained on medical texts in Spanish, thereby offering high efficiency for processing our specific corpora. The resulting bi-encoders are coined as Spanish-SapBERT models.

The bi-encoder models were trained for a single epoch, with a batch size of 256, a learning rate of 2e-5, a maximum input length of 256 tokens, and a hard triple mining margin of 0.2. These parameters were selected to balance the trade-off between training time and model performance. The large batch size was chosen to provide more data for the extraction of negative candidates, enhancing the learning process by presenting a wider variety of contrasts and contexts.  On the other hand, for performance comparison with the Spanish-SapBERT models, we also followed an alternative strategy to train domain-adapted RoBERTa models for each corpora by using input triplets composed as follows: positive terms extracted from the descriptions provided in the gazetteers---supplied together with the DisTEMIST and MedProcNER corpora by the challenge organizers---and a random subset as negative terms (five negative terms per each positive term were chosen). For this corpus-specific bi-encoders, we use triplet-margin-loss implemented in the \textit{Sentence Transformer} library. Finally, to compare the performance of the proposed Spanish-SapBERT bi-encoders with a baseline language model, we have also worked with a domain-adapted language RoBERTa model not trained for MEL tasks, but for other general-purpose tasks.

An overview of the bi-encoder models analyzed in this study is the following:

\begin{itemize}
    \item \textbf{Spanish-SapBERT} variants: these models are based on the \textit{Roberta-base-biomedical-clinical-es}~\cite{carrino2021biomedical} architecture and were trained with Spanish UMLS triplets (+1M medical concepts). The \textit{standard variant} \textit{Spanish-SapBERT} excludes obsolete terms, focusing on current biomedical and clinical contexts in Spanish, while the \textit{obsolete-codes variant} \textit{Spanish-SapBERT-oc} includes obsolete terms to ensure comprehensive coverage of both contemporary and historical medical terminology. This model is publicily available in HuggingFace\footnote{https://huggingface.co/BSC-NLP4BIA/SapBERT-from-roberta-base-biomedical-clinical-es}
    \item \textbf{Corpus-specific} bi-encoders: the \textit{DisTEMIST-biencoder} and \textit{MedProcNER-biencoder}, both grounded in the \textit{Roberta-base-biomedical-clinical-es}~\cite{carrino2021biomedical} model, were trained with corpus-specific triplets. The former focuses on disease entity recognition using DisTEMIST data~\cite{distemist}, whereas the latter focuses on medical procedure identification with data from MedProcNER, showcasing their specialized applications in clinical text analysis.
    \item \textbf{SapBERT-XLMR} multilingual models: they are the \textit{SapBERT-XLMR-base} and \textit{SapBERT-XLMR-large} models~\cite{Liu2021-ik}, leveraging the XLM-Roberta architectures~\cite{Conneau2019-tz}. Both models underwent training with the entire UMLS dataset (+4M medical concepts) and feature cross-lingual adaptation to the languages contained within UMLS, enhancing their global applicability.
    \item \textbf{Roberta-base-biomedical-clinical-es}: this model (abbreviated as \textit{Roberta-base-bio-cli} in the tables an figures below), derived from XLM-Roberta-base~\cite{Conneau2019-tz}, was refined with a comprehensive biomedical corpus in Spanish~\cite{carrino2021biomedical}, setting a robust foundation for adaptation to various clinical tasks.
\end{itemize}

The SapBERT-XLMR bi-encoders were trained on a node with 8 NVIDIA A100 GPUs, while the Spanish-SapBERT variants, the corpus-specific bi-encoders and XLM-Roberta-base models were trained on a single GPU, thus highlighting the improvement in performance and computational requirements for these less complex models.

Additionally, we used the FAISS library~\cite{douze2024faiss} for retrieving the K-nearest neighbors candidates from the embeddings provided by the bi-encoders, enhancing the base results obtained in the candidate-retrieval stage. FAISS is critical for efficiently managing the vast amount of data involved in this process, ensuring large scale-processing. FAISS improves efficiency in both time and space, essential for handling large sets of embeddings and enabling rapid retrieval of nearest neighbors, decisive for our candidate re-ranking process (see equation~\ref{eq:faiss}).

\begin{equation}
\label{eq:faiss}
    \forall M_i \in \{M_1, M_2, \ldots, M_N\}, \quad C_i = \text{FAISS}(\text{bi-encoder}(M_i)), \quad |C_i| = k
\end{equation}

For the candidate re-ranking stage of ClinLinker, we built upon the work done by these same authors for the SympTEMIST-linking subtask~\cite{Gallego2023-qy}, where initial explorations with the Sentence Transformer’s cross-encoder were conducted. This phase is crucial as it refines the initial selection made by the bi-encoder, ensuring that the most relevant candidates are prioritized. After several experiments, we found that the optimal way to train the cross-encoder was using the candidates generated by the bi-encoder, leveraging the similarity between the correct mention and each candidate (see Figure~\ref{fig:bi_cross_pipeline}). This similarity comparison is vital for the cross-encoder to learn the subtle differences between similar candidates, thereby improving its ability to select the most appropriate candidates for a given entity mention. This method is known as training with hard triplets.

For cross-encoder training, we generated candidates from the training set of DisTEMIST and MedProcNER corpora, without using the gazetteer provided by the organizers of those shared tasks, creating the mentioned triplets---\textit{(anchor, positive, negative)}---, always initializing the cross-encoder's parameters with each corresponding bi-encoder's weights. This approach ensures that the cross-encoder is exposed to a wide range of scenarios, including challenging cases where the distinction between candidates is not immediately obvious. At the outset, the effectiveness of fine-tuning domain-adapted models such as \textit{roberta-base-biomedical-clinical-es}~\cite{carrino2021biomedical}---which, unlike SapBERT, do not start from language models pre-trained for the self-alignment of entities---has been also evaluated in this study; however, we noted that this approach did not surpass the performance observed in the bi-encoder+cross-encoder pipelines proposed herein.

The complete ClinLinker's inference process (see Figure~\ref{fig:bi_cross_pipeline} and equations \ref{eq:cross-encoder} and \ref{eq:cross-encoder-rerank}) for each test set mention involves generating candidates with the Spanish-SapBERT bi-encoder, this time using the gazetteer supplied within the given corpora, and obtaining scores for each ($M_i$, $C_i$) pair. By scoring these pairs, we quantitatively assess the relevance of each candidate ($C_i$) to the mention ($M_i$). This allowed us to group them by mentions and sort them according to the obtained scores.

\begin{eqnarray}
\label{eq:cross-encoder}
    \text{Score}_{ij} & = & \text{CrossEncoder}(M_i, C_{ij}) \quad \forall C_{ij} \in C_i \\
\label{eq:cross-encoder-rerank}
    C_i' & = & \text{rerank}(C_i, \text{Score}_{ij}) \quad \forall M_i \in \{M_1, M_2, \ldots, M_N\}
\end{eqnarray}


The final candidate rankings provided by ClinLinker are a mixture of advanced NLP techniques and domain-specific knowledge, ensuring high accuracy in EL across diverse medical texts. All the results obtained with ClinLinker are analyzed in the following section, in which we distinguish between the results obtained for the two corpora, DisTEMIST and MedProcNER. We also separate the analysis of the results obtained on the \textit{gold standard} datasets---which contains all the codes from the provided test sets---from those obtained on the \textit{unseen codes} datasets---encompassing only those codes in the test sets not present in the training sets supplied by the shared-tasks organizers.

\section{Results and Discussion}
\label{sec:results}

To assess the advancements made in this study, two distinct corpora were employed: DisTEMIST~\cite{distemist}, a dataset for automatic detection of disease mentions in clinical cases, and MedProcNER, a dataset for the automated identification of findings and procedures in clinical scenarios. These datasets represent a diverse collection of medical texts, providing a comprehensive testing ground for our models.

The chosen metric for comparative evaluation was top-k accuracy at 25. This metric is particularly relevant for practical applications where a balance between accuracy and a broad range of candidate suggestions is necessary. However, calculations were also performed for the top-k accuracy at 1, 5, 50, and 100. This range of metrics provides a detailed view of the model's performance across different levels of specificity. The selection of these metrics is rooted in the current limitation of models in accurately determining the linkage to a code with a single candidate. Yet, for higher candidate counts, such as @20 (k $= 20$) or @25 (k $= 25$), the predictions exhibit robust performance, demonstrating the model's effectiveness in providing a wide selection of relevant candidates. This robustness is essential in practical scenarios, where identifying an exact match as a unique code among candidates is often challenging due to the presence of cross-synonyms among different codes within the terminology,  a situation that is particularly prevalent in highly granular vocabularies such as SNOMED-CT.

The results achieved are in line with the anticipated ideal behavior of the trained models, indicating the effectiveness of our training approach and the suitability of ClinLinker for practical applications. In both MedProcNER and DisTEMIST datasets, the Spanish-SapBERT bi-encoders, trained exclusively with concept descriptions in Spanish from UMLS, surpass the generic multilingual SapBERT-XLMR base and large models\footnote{\textit{cambridgeltl/SapBERT-UMLS-2020AB-all-lang-from-XLMR-large} and \textit{cambridgeltl/SapBERT-UMLS-2020AB-all-lang-from-XLMR}~\cite{liu2021learning}}, which are currently the benchmarks from the associated shared tasks. Furthermore, the cross-encoder strategy successfully reorders the candidates, providing a refined list where higher-ranked candidates are more likely to be the correct match, thus enhancing the practical utility of the system in clinical settings.

\begin{figure}[!ht]
    \centering
        \includegraphics[width=\linewidth]{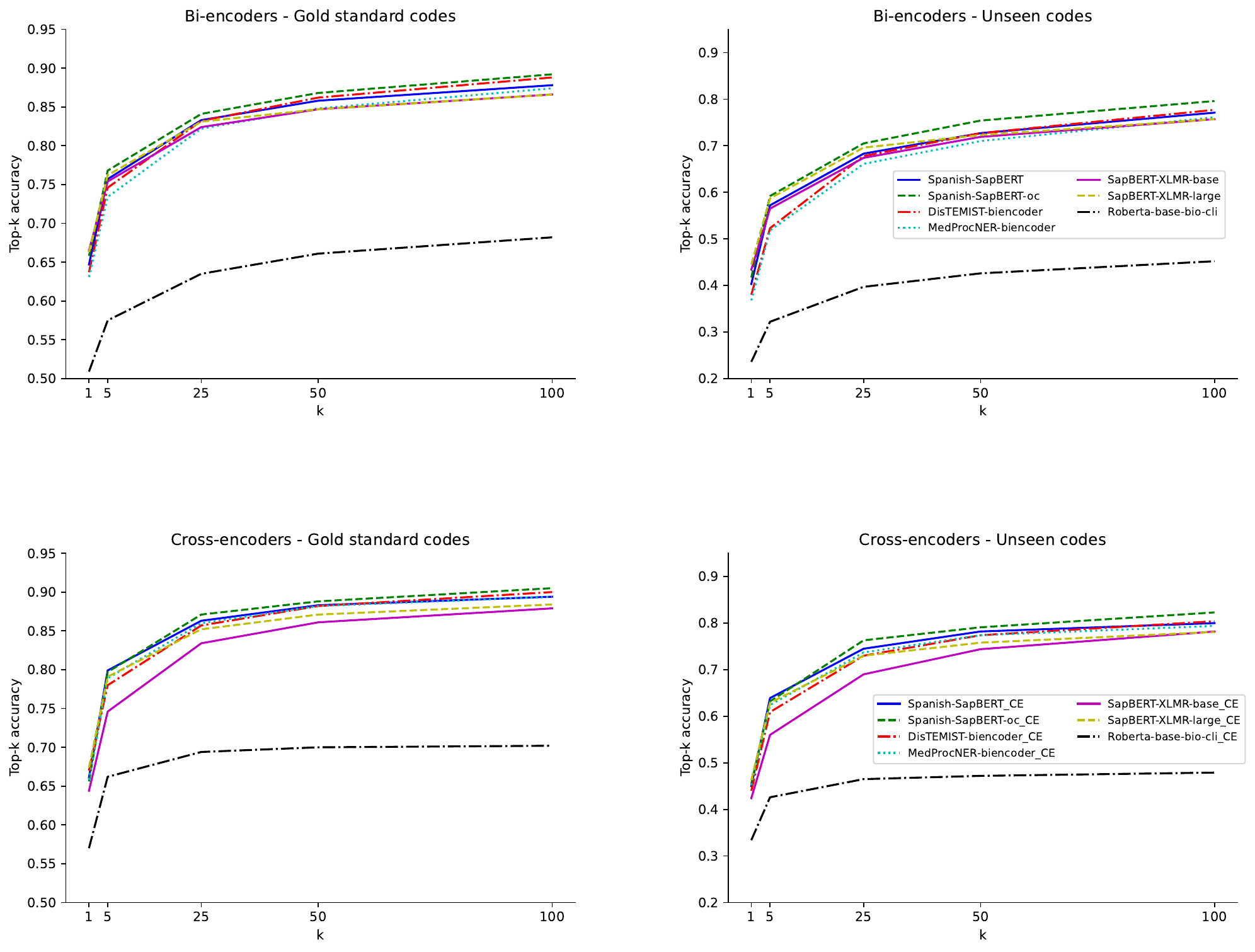}
    \caption{Performance comparison of the bi-encoder and bi-encoder+cross-encoder (``\_CE'') models on the DisTEMIST corpus: efficacy of the models across various retrieval thresholds (top-k accuracy) for both the validated gold-standard annotations and the unseen-codes subsets. (Note that the figures are in different scales to show the differences between the performance of the different models in both subsets.)}
    \label{fig:distemist_results}
\end{figure}

\begin{figure}[!ht]
    \centering
        \includegraphics[width=\linewidth]{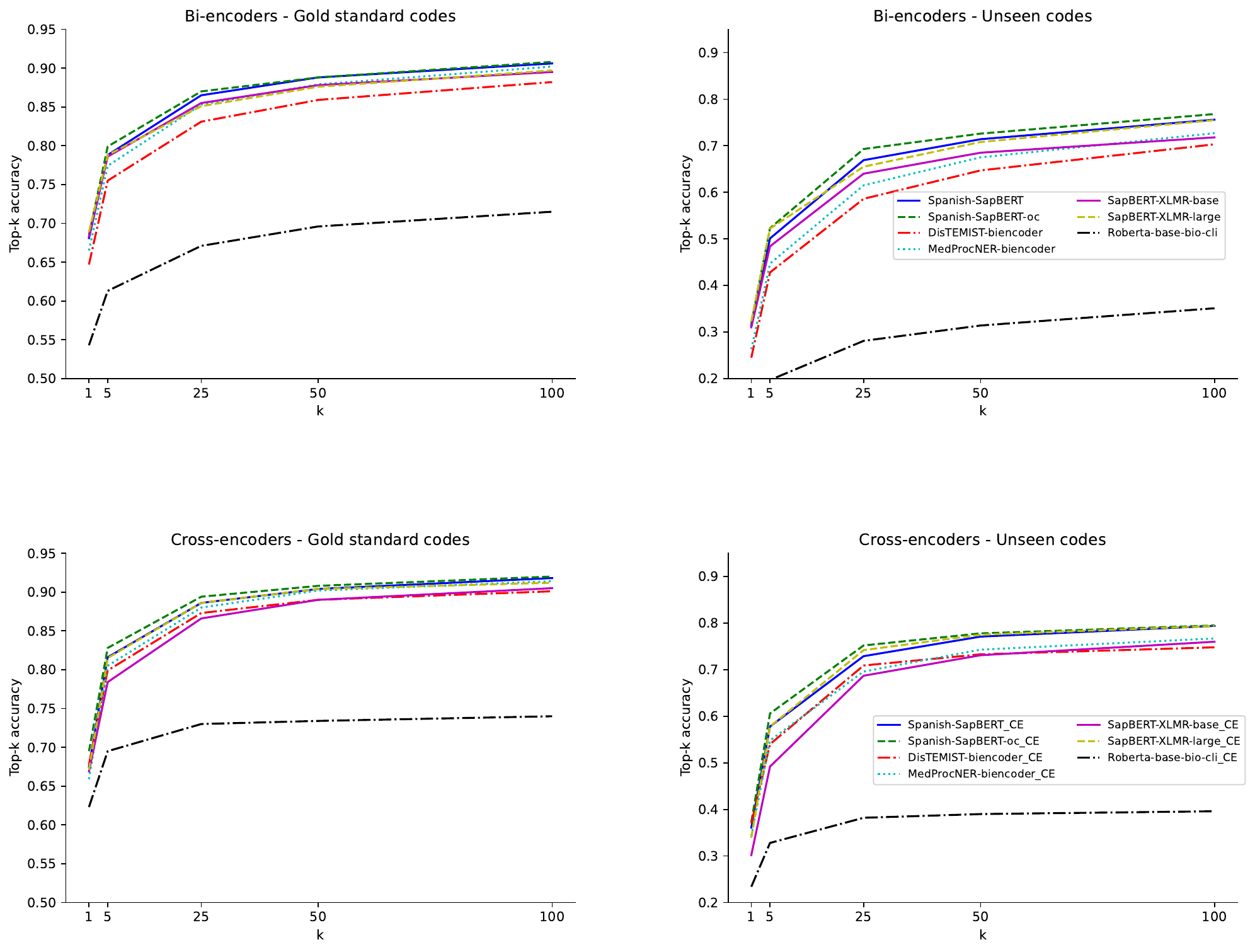}
    \caption{Performance comparison of the bi-encoder and bi-encoder+cross-encoder (``\_CE'') models on the MedProcNER corpus: efficacy of the models across various retrieval thresholds (top-k accuracy) for both the validated gold-standard annotations and the unseen-codes subsets. (Note that the figures are in different scales to show the differences between the performance of the different models in both subsets.)}
    \label{fig:medprocner_results}
\end{figure}

Figures~\ref{fig:distemist_results} and \ref{fig:medprocner_results} show the performance of trained bi-encoder models on the DisTEMIST and MedProcNER corpora. We compare our Spanish-SapBERT bi-encoders used in our ClinLinker pipeline against the multilingual SapBERT-XLMR, highlighting performance improvements, especially with Spanish-SapBERT-oc using outdated terms in the pretraining step. This comparison underscores the superior efficacy of our language-specific bi-encoders over the generic multilingual alternatives.

The difference between the ClinLinker's triplet-based bi-encoder+cross-encoder models (Spanish-SapBERT\_CE and Spanish-SapBERT-oc\_CE) and the current benchmarks---based on the Sapbert-XLMR models---, becomes more pronounced for the DisTEMIST and MedProcNER datasets, as shown in Figures~\ref{fig:distemist_results} and \ref{fig:medprocner_results}, where the ``\_CE'' suffix denotes the complete pipeline, i.e the cross-encoder's re-ranking stage applied to the output candidates of each specific bi-encoder. CinLinker, our two-stage approach for MEL in Spanish, consistently outperforms the SOTA biencoders across both datasets in top-k accuracy at 25 metric. For the DisTEMIST-linking task, we achieve a $4.8\%$ improvement for the gold standard dataset, reaching $0.871$, and a $9.6\%$ increase for the zero-shot scenario, reaching a $0.763$, surpassing all top-k accuracy values except for top-1 accuracy. Similarly, for the MedProcNER corpus, our approach shows a similar trend, slightly outperforming models used in SOTA. We improve top-k accuracy at 25 by $5.1\%$ for the gold-standard dataset and $14.8\%$ for the unseen-codes dataset, achieving noteworthy values of $0.894$ and $0.752$, respectively.

Tables \ref{tab:distemist} and \ref{tab:medprocner} summarize the performance results for ClinLinker and all the analyzed bi-encoder and bi-encoder+cross-encoder models for MEL on the DisTEMIST and MedProcNER corpora, respectively.

\begin{table}
\centering
\caption{Comparative results (top-k accuracy) of selected bi-encoder (last 7 rows) and bi-encoder+cross-encoder (``\_CE'', first 7 rows) models on the DisTEMIST \textit{gold standard} and \textit{unseen codes} dataset. ($\dag$: ClinLinker's bi-encoder alone; $\ddag$: ClinLinker's bi-encoder+cross-encoder. \textbf{Bold}: best result for each top-k accuracy; \underline{underlined}: second best; \textit{italic}: current benchmark).}
\label{tab:distemist}
\begin{tabular}{lccccc|ccccc}
\hline
& \multicolumn{5}{c|}{gold standard} & \multicolumn{5}{c}{unseen codes}\\ \hline
\textbf{MODEL'S NAME} & \textbf{@1} & \textbf{@5} & \textbf{@25} & \textbf{@50} & \textbf{@100}  & \textbf{@1} & \textbf{@5} & \textbf{@25} & \textbf{@50} & \textbf{@100}   \\ \hline
Spanish-SapBERT\_CE$^\ddag$ & .661 & \textbf{.799} & \underline{.863} & \underline{.883} & .894  & .450 & \textbf{.639} & \underline{.745} & \underline{.782} & .800  \\
Spanish-SapBERT-oc\_CE$^\ddag$ & .656 & \underline{.797} & \textbf{.871} & \textbf{.888} & \textbf{.905}  & \underline{.455} & \underline{.632} & \textbf{.763} & \textbf{.791} & \textbf{.823}  \\
DisTEMIST-biencoder\_CE & \underline{.670} & .780 & .857 & .882 & \underline{.900}  & .440 & .609 & .730 & .774 & \underline{.804} \\
MedProcNER-biencoder\_CE & .658 & .789 & .860 & .882 & .894 & .452 & .624 & .737 & .774 & .794  \\
SapBERT-XLMR-base\_CE & .644 & .746 & .834 & .861 & .879 & .424 & .560 & .690 & .744 & .782 \\
SapBERT-XLMR-large\_CE & \textbf{.674} & .791 & .852 & .871 & .884 & \textbf{.463} & .629 & .730 & .758 & .781 \\
Roberta-base-bio-cli\_CE & .570 & .662 & .694 & .700 & .702 & .334 & .426 & .465 & .472 & .479 \\
\cdashline{1-11}
Spanish-SapBERT$^\dag$ & .647 & .757 & .833 & .858 & .878  & .403 & .572 & .683 & .727 & .771  \\
Spanish-SapBERT-oc$^\dag$ & .658 & .768 & .841 & .868 & .892  & .417 & .592 & .705 & .754 & .796  \\
DisTEMIST-biencoder & .637 & .746 & .832 & .862 & .888  & .380 & .523 & .678 & .727 & .777 \\
MedProcNER-biencoder & .631 & .734 & .822 & .848 & .874  & .368 & .518 & .661 & .710 & .761  \\
SapBERT-XLMR-base & .665 & .754 & .824 & .847 & .866 & .434 & .565 & .674 & .719 & .757 \\
\textit{SapBERT-XLMR-large} & \textit{.663} & \textit{.762} & \textit{.831} & \textit{.847} & \textit{.866} & \textit{.445} & \textit{.587} & \textit{.696} & \textit{.723} & \textit{.757}  \\
Roberta-base-bio-cli & .509 & .575 & .635 & .661 & .682 & .236 & .322 & .397 & .426 & .452 \\
\hline
\end{tabular}
\end{table}

\begin{table}
\centering
\caption{Comparative results (top-k accuracy) of selected bi-encoder (last 7 rows) and bi-encoder+cross-encoder (``\_CE'', first 7 rows) models on the MedProcNER \textit{gold standard} and \textit{unseen codes} dataset. ($\dag$: ClinLinker's bi-encoder alone; $\ddag$: ClinLinker's bi-encoder+cross-encoder. \textbf{Bold}: best result for each top-k accuracy; \underline{underlined}: second best; \textit{italic}: current benchmark).}
\label{tab:medprocner}
\begin{tabular}{lccccc|ccccc}
\hline
& \multicolumn{5}{c|}{gold standard} & \multicolumn{5}{c}{unseen codes}\\ \hline
\textbf{MODEL'S NAME} & \textbf{@1} & \textbf{@5} & \textbf{@25} & \textbf{@50} & \textbf{@100} & \textbf{@1} & \textbf{@5} & \textbf{@25} & \textbf{@50} & \textbf{@100}  \\ \hline
Spanish-SapBERT\_CE$^\ddag$ & .675 & \underline{.816} & \underline{.886} & \underline{.904} & \underline{.918} & .361 & \underline{.578} & .729 & .771 & \underline{.794} \\
Spanish-SapBERT-oc\_CE$^\ddag$ & \textbf{.695} & \textbf{.828} & \textbf{.894} & \textbf{.908} & \textbf{.920} & \textbf{.373} & \textbf{.606} & \textbf{.752} & \textbf{.778} & \textbf{.795} \\
DisTEMIST-biencoder\_CE & .678 & .799 & .873 & .890 & .901 & \underline{.370} & .540 & .709 & .733 & .748 \\
MedProcNER-biencoder\_CE & .659 & .805 & .880 & .902 & .914 & .343 & .548 & .696 & .743 & .767 \\
SapBERT-XLMR-base\_CE & .669 & .784 & .866 & .890 & .905 & .302 & .492 & .687 & .731 & .760 \\
SapBERT-XLMR-large\_CE & .671 & .815 & \underline{.886} & \underline{.904} & .912 & .339 & \underline{.578} & \underline{.742} & \underline{.775} & \underline{.794} \\
Roberta-base-bio-cli\_CE & .623 & .695 & .730 & .734 & .740 & .234 & .328 & .382 & .390 & .396 \\
\cdashline{1-11}
Spanish-SapBERT$^\dag$ & .681 & .788 & .865 & .888 & .906 & .314 & .501 & .669 & .714 & .756 \\
Spanish-SapBERT-oc$^\dag$ & .685 & .799 & .870 & .888 & .908 & .318 & .523 & .693 & .726 & .768 \\
DisTEMIST-biencoder & .647 & .755 & .831 & .859 & .882 & .245 & .428 & .586 & .647 & .703 \\
MedProcNER-biencoder & .665 & .774 & .853 & .879 & .902 & .263 & .447 & .615 & .675 & .727 \\
SapBERT-XLMR-base & .683 & .786 & .855 & .878 & .895 & .310 & .484 & .640 & .685 & .718 \\
\textit{SapBERT-XLMR-large} & \textit{\underline{.689}} & \textit{.788} & \textit{.851} & \textit{.876} & \textit{.897} & \textit{.323} & \textit{.522} & \textit{.655} & \textit{.708} & \textit{.755} \\
Roberta-base-bio-cli & .543 & .613 & .671 & .696 & .715 & .133 & .197 & .281 & .314 & .351 \\
\hline
\end{tabular}
\end{table}

Results clearly demonstrate that ClinLinker, specifically tailored to the Spanish language, significantly outperforms the multilingual onset, despite being trained on a subset of the data used to train the multilingual XLM-Roberta based models. This superiority underscores the critical importance of linguistic adaptation in model training, particularly when dealing with complex clinical data. Furthermore, the effectiveness of our model is not confined to its linguistic specificity; it is sufficiently versatile to surpass the SOTA in two different corpora annotated with SNOMED-CT codes. This finding emphasizes the broad applicability and efficacy of ClinLinker, highlighting its potential utility across a diverse range of clinical and linguistic contexts.

These advancements are not just incremental but represent a considerable leap forward in the field of NLP applied to clinical text exploitation. The improvement in unseen codes is particularly noteworthy as it indicates the robustness and generalizability of our model, fundamental for real-world clinical applications where unpredictability is the norm. This success can be attributed to the focused approach of our model, which, by prioritizing linguistic specificity, demonstrates the profound impact of language-tailored solutions in the domain of clinical informatics. Additionally, the improvement in gold standard performance suggests that our model is highly effective in identifying correct codes, even among a vast array of possibilities, which is paramount for practical clinical usage.

\section{Conclusions}
Our study marks a significant leap in NLP applied to medical text analysis. Utilizing a two-phase approach with a bi-encoder for candidate retrieval, followed by a cross-encoder for candidate re-ranking, with models trained with contrastive-learning strategies and specifically tailored for the Spanish language, ClinLinker has demonstrated superior performance compared to existing multilingual models for MEL tasks. Moreover, our models trained with the entire set of terms in Spanish from the UMLS ontology show superior performance rates than those ones trained with the specific corpora, proving to facilitate their reuse for new corpora tagged with UMLS subsets such as SNOMED-CT. Also, the results presented in this paper underscore the high impact of our models in clinical settings, offering high accuracy for both  top-k accuracy at 5 and top-k accuracy at 25 metrics. This level of precision is more than acceptable for semi-automatic annotation processes, underscoring the practical applicability of our approach in real-world clinical environments. These findings not only highlight the importance of linguistic adaptation in model training but also underscore the potential of our approach in enhancing the utility of digital medical records across various clinical contexts. The robustness and generalizability of our MEL models, particularly in handling unfamiliar codes, pave the way for future research focused on language-specific NLP solutions in clinical informatics. The presented approach showed promising results and could be adapted also for concept normalization of other types of clinical relevant entities such as drugs, chemical compounds and proteins or for clinical variable extraction and data structuring applied to more tailored use cases for instance in cardiology or other medical specialities.

For future research, we aim to work not only with mentions but also with the context of these mentions, endowing MEL language models with information usable for both generating and ordering candidate medical concepts to be matched to the target mentions. Additionally, we plan also to further explore the use of knowledge-graph (KG) enhanced language models, by leveraging KGs obtained from UMLS or SNOMED-CT to enrich MEL models with in-domain semantic information.



%
%
%
%
%
\bibliographystyle{splncs04}
\bibliography{gallego_et_al_iccs2024}
\end{document}